\pdfoutput=1
\documentclass[letterpaper, 10 pt, conference]{ieeeconf}  

\IEEEoverridecommandlockouts                              

\usepackage{graphics} 
\usepackage{epsfig} 
\usepackage{subfigure} 
\usepackage{url} 
\usepackage{amsmath} 
\usepackage{amssymb}  
\usepackage{mathtools}

\usepackage[table]{xcolor}

\usepackage{xcolor,colortbl}
\definecolor{green2}{rgb}{0.1,0.5,0.1}
\definecolor{blue2}{rgb}{0.,0.75,0.75}

\newcommand\kevinupdate[1]{\textcolor{black}{#1}}

\title{\LARGE \bf
Single-Shot Clothing Category Recognition in Free-Configurations with Application to Autonomous Clothes Sorting}

\author{Li Sun$^{1,2}$, Gerardo Aragon-Camarasa$^{1}$, Simon Rogers$^{1}$, Rustam Stolkin$^{2}$, J. Paul Siebert$^{1}$
\thanks{This work was supported by: European FP7 Strategic Research Project CloPeMa, www.clopema.eu; H2020 RoMaNs, 645582, www.h2020romans.eu; EPSRC grant EP/M026477/1.}
\thanks{$^{1}$School of Computing Science, University of Glasgow, G12 8QQ, Glasgow, UK
        {\tt\small l.sun.1@research.gla.ac.uk}}
\thanks{$^{2}$Extreme Robotics Lab, University of Birmingham, Birmingham, B15 2TT, UK.
        {\tt\small lisunsir@gmail.com@gmail.com}}%
}

\begin{document}

\maketitle
\thispagestyle{empty}
\pagestyle{empty}

\begin{abstract}
This paper proposes a single-shot approach for recognising clothing categories from 2.5D features.  We propose two visual features, BSP (B-Spline Patch) and TSD (Topology Spatial Distances) for this task. The local BSP features are encoded by LLC (Locality-constrained Linear Coding) and fused with three different global features. Our visual feature is robust to deformable shapes and our approach is able to recognise the category of unknown clothing in unconstrained and random configurations. We integrated the category recognition pipeline with a stereo vision system, clothing instance detection, and dual-arm manipulators to achieve an autonomous sorting \kevinupdate{system}. To verify the performance of our proposed method, we build a high-resolution RGBD clothing dataset of 50 clothing items of 5 categories sampled in random configurations (a total of 2,100 clothing samples). Experimental results show that our approach is able to reach 83.2\% accuracy while classifying clothing items which were previously unseen during training. This advances beyond the previous state-of-the-art by 36.2\%. Finally, we evaluate the proposed approach in an autonomous robot sorting system, in which the robot recognises a clothing item from an unconstrained pile, grasps it, and sorts it into a box according to its category. Our proposed sorting system achieves reasonable sorting success rates with single-shot perception.
\end{abstract}

\section{INTRODUCTION}
Recognising, understanding and handling of textiles and clothing is an example of a task which seems very simple to humans, but which poses profound challenges to robotics and AI, demanding a close integration of perception, action, learning, and reasoning. While such challenges are of great academic interest, there is also significant and urgent demand in economically large and societally important industries, which might not be obvious to the layperson. For example, robots are needed for sorting and segregation of huge quantities of hazardous nuclear waste, which can contain numerous kinds of complex deformable materials (contaminated protective suits, overalls, gloves, tarpaulins), and where recognition of individual objects is extremely important for judging the level of hazard (e.g. when sorting and inventorying the unknown contents of half-century old waste containers) \cite{marturi2016towards}.

\begin{figure*}[t]
\centering
\includegraphics[width = 0.98\textwidth]{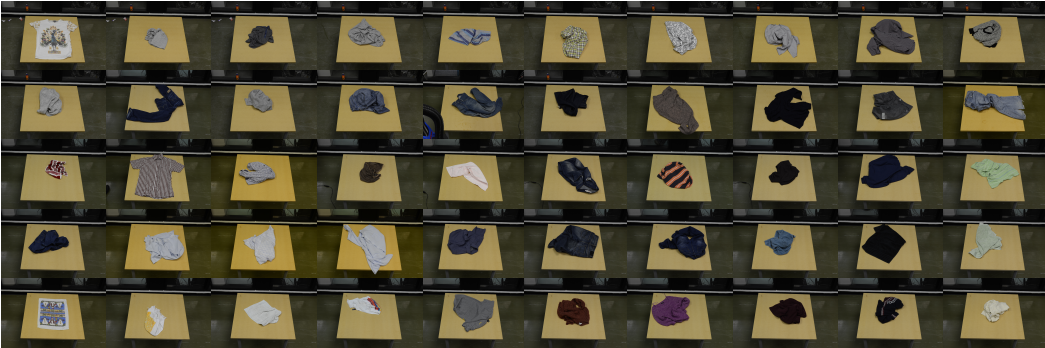}
\caption{A subset of clothing items in our dataset. In our dataset, there are 50 clothing items of 5 categories of different shapes and colours. This dataset can be downloaded at {https://sites.google.com/site/clopemaclothesdataset/}.}
\label{fig:dataset}
\end{figure*}

We propose a novel robot vision recognition pipeline for classifying items of clothing, based on features which represent and describe material attributes. Our approach is able to recognise clothing categories in arbitrary (crumpled) configurations (as shown in Fig.~\ref{fig:dataset}). This task is extremely challenging for two reasons. Firstly, clothing in free configurations can be highly wrinkled and self-occluding, hence it is difficult to encode clothing categories into generic visual representations. Secondly, because clothes are highly deformable objects, which have almost infinite possible configurations, learning a generalised category model from limited training data is difficult.

To overcome the above challenges, we propose a generic approach to recognise clothing categories from 2.5D surfaces. Instead of detecting the components of clothing such as collars, sleeves, pockets, etc., or describing clothing patterns using RGB-based features, we focus on material attributes -- fabric, thickness and stiffness. For instance, an item of clothing is more likely to be a sweater if it is knitted, to be a shirt if made of jaconet, and to be jeans if made of denim. 

The contributions of this paper are: (1), we describe a novel approach for recognising clothing categories in random configurations based on high-quality stereo data. Experimental results show that our approach advances beyond the previous state-of-the-art \cite{ramisa2013finddd} with 36.2\% improvement in classification accuracy; (2), our approach generalises and is robust to configurations and categories of clothing which were not seen during training; (3), a generic robot vision approach for recognising clothing categories is presented and integrated within an autonomous sorting pipeline using a dual-arm robot. 

\section{Related Work\label{sec:rw}}
Existing clothing category recognition systems can be divided into two groups. Those based on perceiving the state of clothing as observed from RGBD data \cite{willimon2011classification,willimon2013classification} and those based on physical simulated models \cite{kita2002model,kita2009clothes,kita2009method,li2014recognition,li2014real}. 

Approaches based on the former include Willimon, et. al. \cite{willimon2011classification} who proposed an interactive pipeline to recognise clothing categories from hanging poses by acquiring multiple views of each item while a robot-arm rotates the item. Visual features used for recognition are based on binary silhouettes and clothing edges. Nearest neighbour classification in feature space is used for categorisation. Thereafter, Willimon, et. al. \cite{willimon2013classification} proposed a mid-level representation, in which 17 mid-level semantic classifiers of clothes attributes (e.g. collar, buttons, hem, colours, patterns, etc.) are trained from low-level RGBD features. SIFT~\cite{sift2004ijcv}, Fast Point Feature Histogram (FPFH)~\cite{fpfh2009icra} and Bag of Features (BoF) coding\cite{bof2006} combined with global features are used as the low-level representation while the output of the 17 mid-level semantic classifiers encode a binary representation of the garment. In this paper, however, we focus on material types and generic 2.5D surfaces. We do not consider semantic components since they are susceptible to occlusion. We also ignore RGB information since it is not a stable representation for \kevinupdate{small-scale clothing categorization problems. Therefore,} RGB data requires a sheer number of training examples to obtain classification results above chance \kevinupdate{i.e. deep learning approaches. That is our approach is suitable when training data is scarce.}

Similarly, Ramisa, et. al. \cite{ramisa2013finddd} devised a 2.5D local descriptor, FINDDD by constructing surface normal histograms over quasi-equidistant bins in the Euclidean space. They used FINDDD together with Bag of Features (BoF) for clothes category recognition in free-configurations. Our approach, however, differs on the robustness of intra-class dissimilarity. For example, they used one polo-shirt to represent the category of polo-shirts. For clothes category recognition using depth data, the extra-class similarities are much larger than intra-class similarities, however, there is no guarantee that intra-class similarities can be neglected. Likewise, their experimental results consist of recognising unknown configurations of known clothes. Dividing training and testing sets on configuration-level instead of clothing-level opens up the possibility of over-fitting. 

Approaches based on simulated models include Li et. al~\cite{li2014recognition} where they proposed a supervised learning approach for recognising clothing categories in hanging poses. They used dense SIFT and sparse coding~\cite{lee2006efficient} as the underlying visual representation. They then optimised their representation by using a binary volumetric representation to achieve real-time performance~\cite{li2014real}. Physical simulated models benefit from the ability to generate the necessary training data for training a classification model. However, clothing items of free configurations are extremely difficult to simulate, e.g. \cite{wong2014improvements} reports the difficulties in simulating: (a) particle inter-collisions between clothing surfaces; and, (b) the static and dynamic frictions between garments and a supporting surface.


Physical simulated studies \cite{kita2009clothes,li2014recognition,li2014real} mainly report recognition of garments in hanging configurations, since the configuration space is greatly reduced. However, a large robot with a large working space is required. Hence, medium sized robots cannot be used to manipulate adult garments. As free-configuration clothing typically presents several occlusions, and a much larger configuration space, the performance of existing simulated approaches within these scenarios is limited. Compared to the widely-used silhouette, RGB patterns and component features, 2.5D clothing material types provide, as presented in this paper, generic and invariant material attributes which are robust to occlusions and random pose configurations. As describe in the following sections, we therefore propose a pipeline incorporating a rich visual description of clothing material type attributes.

In our previous research\cite{sun2016icra}, three types of global features (i.e. LBP, Shape Index and a preliminary version of TSD) are used for clothing category recognition. In this work, a single-shot perception cannot yield satisfactory performance, thereby a GP-based interactive perception strategy is designed to reconfigure the highly-wrinkled clothing to recognisable configurations. In this paper, we aim to advance the single-shot recognition performance by enriching the robust visual representations. 


\section{Clothes Category \& Material Recognition}
\subsection{Outline\label{sec:outline}}

\begin{figure*}[th]
\centering
\includegraphics[width= 0.98\textwidth]{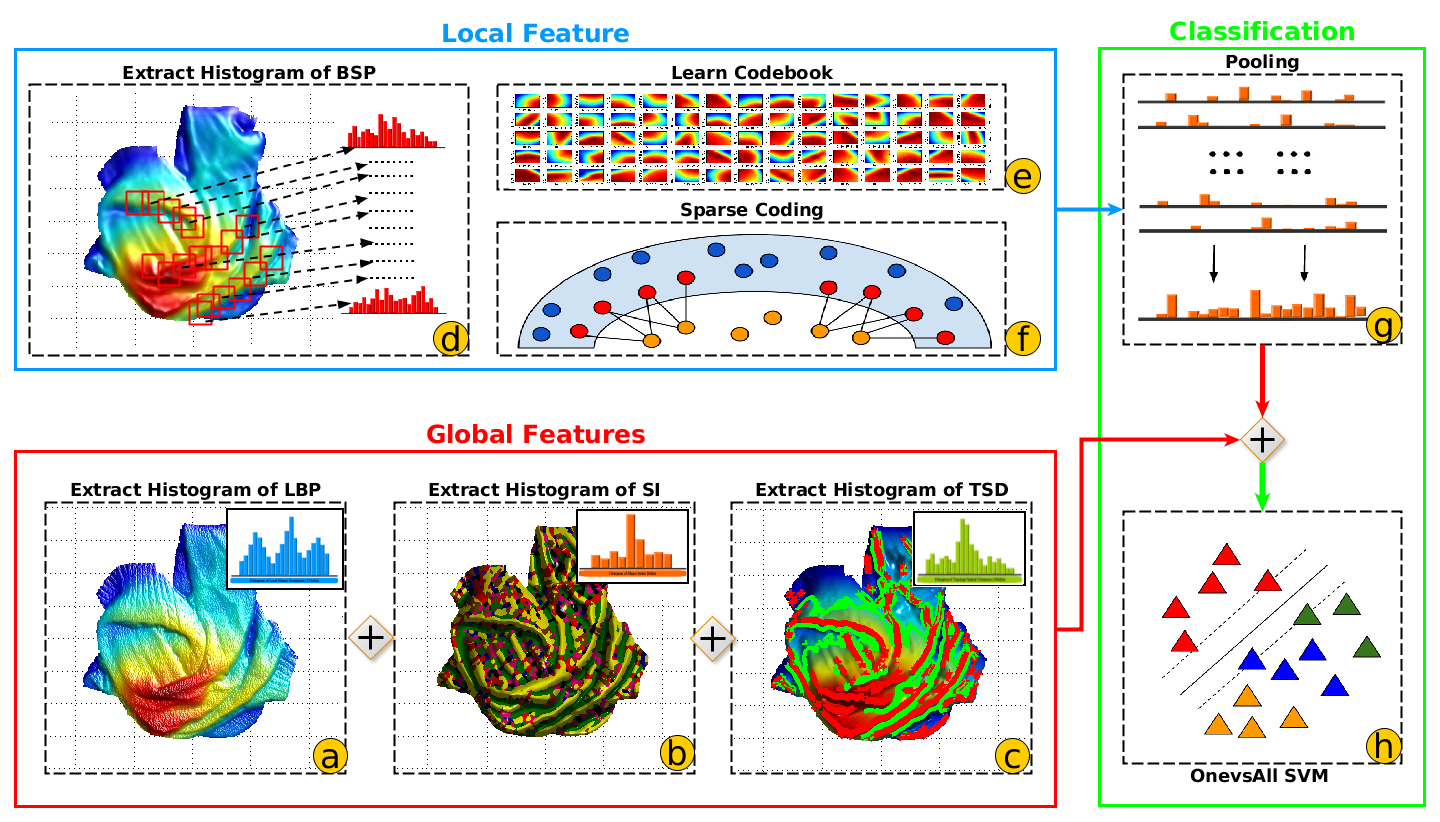}
\caption{The proposed pipeline for clothing category recognition. \kevinupdate{The pipeline has three phases: local feature extraction (with global coding), global features extraction and classification.}}
\label{fig:pipeline}
\end{figure*}

Our robot vision recognition pipeline (Fig. \ref{fig:pipeline}) consists of four modules: (1) global and (2) local feature extraction, (3) encoding, and (4) classification. For global features, Shape Index (SI) and Local Binary Pattern (LBP) descriptors are extracted (Section{~\ref{sec:si}} and Section{~\ref{sec:lbp}}, respectively). We also propose a Topology Spatial Distance (TSD) feature computed from surface topologies (Section{~\ref{sec:tsd}}), which captures the intrinsic properties of wrinkles. For local features, $B$-Spline Patches (BSP) (Section{~\ref{sec:bsp}) are extracted densely on surface ridges and then encoded using sparse coding in order to represent the 2.5D shape of wrinkles. Finally, global and local descriptions are fused together and fed into a classifier in order to learn clothing category models (Section{~\ref{sec:svm}}). 

2.5D Surface textures of clothing are useful for recognising materials types. Among texture recognition approaches, LBP have achieved great success in grey scale texture recognition \cite{lbpPAMI2002}. In this paper, LBP is applied to the high frequency phase of 2.5D clothes surfaces to describe the 2.5D fabric structure. Wrinkles provide another important feature of clothing material, specially in free-configuration settings. Wrinkles can reveal thickness and stiffness attributes of the fabric material. For instance, wrinkles in t-shirts are usually smaller than those of sweaters because of their thickness attributes, whilst jeans usually have fewer wrinkles than other clothing types since denim is stiffer. 

\subsection{A Generic Clothing Surface Analysis Framework}

In our previous work~\cite{lisun2015icra}, we reported a generic clothing surface analysis framework for garment grasping and flattening in robotic tasks. In this paper, we extend this framework as presented in Section~\ref{sec:si}, Section~\ref{sec:tsd} and Section~\ref{sec:bsp}. For completeness, a brief description of our previous framework is given below.

A piece-wise $B$-Spline surface approximation is adapted to fit a continuous implicit surface onto the original depth map (Fig. \ref{fig:pipeline}-d). Surface shape and topology features are derived from low-level surface geometries such as curvatures. To detect wrinkles on the garment surface, we compute the surface's topology which includes surface ridges and wrinkle contours (shown as red and green lines in Fig.~\ref{fig:pipeline}-c). We detect ridges by thresholding the maximal curvature at different scales, and the preliminary result is filtered by `ridge' regions (Section~\ref{sec:si}). Wrinkle contours are estimated by computing the zero-crossings of the second derivatives of the garment's surface (further details can be found in our previous work \cite{lisun2015icra}). Morphological image operations~\cite{lam1992thinning} are applied to thin ridges and wrinkle contours to obtain ridges of one-pixel-width.

\subsection{Global Features\label{sec:global}}
\subsubsection{Histogram of Shape Index (SI)\label{sec:si}}
Shape Index~\cite{koenderink1992surface} classifies surface regions into real-valued index values $S$ in the range $[-1,1]$. The SI value is quantised into 9 intervals corresponding to 9 surface types -- cup, trough, rut, saddle rut, saddle, saddle ridge, ridge, dome and cap. Amongst shape types, `ridge' (shown as yellow in Fig.~\ref{fig:pipeline}-b) is critical in the analysis and description of wrinkles. The shape index value $S^{p}$ of point $p$ can therefore be calculated as follows~\cite{koenderink1992surface}:
\begin{equation}
\centering
\displaystyle S^{p}= \frac{2}{\pi} \tan^{-1} \left [ \frac {k_{min}^{p}+k_{max}^{p}}{k_{min}^{p}-k_{max}^{p}} \right ],
\label{eq:12}
\end{equation}
where $k_{min}^{p}$ and $k_{max}^{p}$ are the minimal and maximal curvatures at point $p$, respectively. In order to parse shape information exhibited by the visible cloth surface, we calculate the shape index from the $B$-Spline fitted depth map and apply majority rank filtering (Fig. \ref{fig:pipeline}-b). This non-linear filtering removes outlier surface classifications and can be tuned to produce a relatively clean classification of shape types over the cloth surface. Finally, a 9 dimensional histogram of shape indices is constructed and then $L^{2}$ normalisation is applied.

\subsubsection{Histogram of Local Binary Patterns (LBP)\label{sec:lbp}}
We extract local binary patterns over multiple-scales in the raw depth map of the visible clothing surface. All descriptors are quantified into a global histogram. That is, LBP histograms are calculated separately at different scales. In our implementation, a selected collection of patterns (58 patterns) are used~\cite{vedaldi08vlfeat}. For multiple-scale feature extraction, a 3 layered Gaussian pyramid is constructed using a sub-division factor of 2 and a Gaussian smoothing parameter ($\sigma$ = 0.375). A global LBP descriptor of 174 dimensions (58$\times$3) is obtained.


\subsubsection{Histogram of Topology Spatial Distances (TSD)\label{sec:tsd}}
Having obtained surface topologies (Section~\ref{sec:si}), we estimate wrinkles' magnitude by calculating the distance between each ridge point and its nearest contour points. Given all surface ridge points $R=\{r_{1},...,r_{n_{r}}\}$ and each wrinkle's contour points $W=\{w_{1},...,w_{n_{w}}\}$, the TSD along the $x-y$ plane \kevinupdate{(w.r.t the camera plane)}, $TSD_{i}^{xy}$, and depth direction, $TSD_{i}^{d}$, are calculated as follows: 
\begin{equation}
\centering
TSD_{i}^{xy} = \min_{i\in n_w} \Vert r_{i},w_{j} \Vert^{2}
\end{equation}
\begin{equation}
\centering
TSD_{i}^{d} = d^{r}_{i} - d^{c}_{\arg_{j} \min \Vert r_{i},w_{j} \Vert^{2}},
\end{equation}
where $d^{r}_{i}$ and $d^{w}_{j}$ are the depth value of the $r_{i}$ and $w_{j}$. The TSD description is a bi-dimensional histogram of topological spatial distances $\{TSD_{1}^{xy},...,TSD_{n_{r}}^{xy}\}$ and $\{TSD_{1}^{d},...,TSD_{n_{r}}^{d}\}$, which represents the wrinkles' estimated width and height, respectively. The final TSD description is obtained by vectorising the bi-dimensional histogram. Width values smaller than 5 or larger than 50 are removed. For both $TSD^{xy}$ and $TSD^{d}$ dimensions, bins are set as 10 uniform intervals ranging from 5 to 50 (the unit on x-y plane is pixel, and on the depth axis is millimetre). The latter corresponds to the possible range of wrinkle widths and heights -- for $TSD^{xy}$, the unit is pixels, while for $TSD^{d}$ the unit is millimetres. After applying $L^{2}$ normalisation, the final 100 dimensional TSD global descriptor (10$\times$10) is obtained.

\subsection{Local Features}
\subsubsection{Local $B$-Spline Patch (BSP)\label{sec:bsp}}
$B$-Spline surface is a classic 3D surface representation in computer graphics, where the surface shape can be manipulated by a set of control points. In our approach, we adapt $B$-Spline surface fitting to describe local clothing patches. That is, for each $m\times n$ patch $P$ of the depth map, a set of 3D points are given ${X\{x_{1},...,x_{n}\}}$,${Y\{y_{1},...,y_{m}\}}$, $P=\{P_{(x_{1},y_{1})},...,P_{(x_{i},y_{j})},...,P_{(x_{n},y_{m})}\}$, where $x_{i}$,$y_{i}$ are x, y coordinates and $P_{(x_{i},y_{i})}$ is the value of the depth map. The implicit surface can be represented as:
\begin{equation}
P(u,w) = \sum_{i=1}^{n+1}\sum_{j=1}^{m+1} \Omega_{i,j}\alpha_{i,k}(x)\beta_{j,l}(y),
\label{eq:1}
\end{equation}
where $\Omega_{i,j}$ represents the control point at row $i$ and column $j$. $\alpha_{i,k}(x)$ and $\beta_{j,l}(y)$ are the basis functions in the $x-y$ plane, more details of which can be found in~\cite{rogers2001introduction}. Eq.\ref{eq:1} can be written in matrix form as:
\begin{equation}
[P]=[\Phi][\Omega],
\label{eq:4}
\end{equation}
where $\Phi_{i,j}=\alpha_{i,k}\beta_{j,l}$, while $P$ is an $r\cdot s\times 3$ matrix containing the 3D coordinates of the range map points, $\Phi$ is a $r\cdot s\times n\cdot m$ basis function matrix containing all the products of $\alpha_{i,k}(u)$ and $\beta_{j,l}(w)$, and $\Omega$ is a $n\cdot m \times 3$ matrix of control point coordinates. Thus, the $B$-Spline surface approximation is obtained by solving {Eq.~\ref{eq:4}} as a least-square problem.

Finally, we use control points $\Omega$ as the local surface representation which is a subset representation of the total set of surface points $P$. In our experiments, a 3rd order uniform open knot vector $[0~0~0~0~1~2~2~2~2]$ is used to compute the bases functions. Each patch is represented by $5\times 5$ control points. Since the control points are distributed uniformly in the $x-y$ plane, only 25 depth values are used for the descriptor. In our approach, the BSP descriptors are extracted densely \kevinupdate{from $m$ $\times$ $n$ patches\footnote{In our implementation, we set $m=n=$35 depending on the practical experience.}} on wrinkles (ridges), instead of extracting them uniformly across the clothing, since our objective is to describe the 3D shape of wrinkles (Section{~\ref{sec:si}}).



\subsubsection{Locality-Constrained Linear Coding}
The Bag-of-Features (BoF) technique only projects a descriptor to its nearest atom in the codebook. Sparse coding allows the local descriptor to be represented by more than one codebook atom. Compared to traditional $L^{1}$-norm sparse coding~\cite{lee2006efficient}, Locality-constrained Linear Coding (LLC)~\cite{wang2010locality} adopts a locality based constraint to enforce sparsity which has been shown to perform more effectively and efficiently in object recognition benchmarks. In this paper, we modified the LLC loss function as:
\begin{equation}
\centering
\begin{split}
\min_{C} \sum_{i=1}^{N}\Vert x_{i}-c_{i} B \Vert^{2} + \lambda \Vert d_{i} \odot \omega \odot c_{i} \Vert ^2 \\
s.t.~  c\textbf{1} = 1, w\textbf{1} = 1, \forall i ~~~~~~~~
\end{split}
\end{equation}
where, N is the number of descriptors, $B^{K\times D}$ is the codebook (generated by $K$-means clustering, $D$ is the dimension of BSP descriptor), $c_{i}^{1\times K}$ is the code for the $i$th descriptor, $x_{i}^{1\times D}$.  $\odot$ refers to element-wise multiplication. $d_{i}$ is the Euclidean distance between $x_{i}$ and the codebook atoms, $\omega=\{\omega_{1},...,\omega_{K}\}$ is the weight of the atoms, calculated by: 

\begin{equation}
\centering
\omega_{j} = \frac{1}{1+e^{-\sigma (n_{j}-\bar{n})}}, 
\end{equation}
where $n_{j}$ is the number of descriptors assigned to the $j$th $K$means cluster, and $\bar{n}$ is $N/K$. In our implementation, $\sigma$ is set to $0.5*10^{-2}$ based on practical experience. The motivation of weighting codebook atoms is based on the assumption that patterns from smaller clusters are more distinctive than those from large clusters. The weights of atoms are set as the sigmoid function of the size of corresponding clusters. 

LLC retrieves a very small number $k$($\ll$ $K$)\footnote{From empirical validation, we found that a $k$ value of 5 works well in practice.} of codebook atoms which are relevant to the matched descriptor. LLC then generates codes in the local coordinate from the retrieved codebook atoms. Sum-pooling is used to generate the global representation:
\begin{equation}
sumpooling:~c_{out} = sum (c_{1},...,c_{n})
\end{equation}
where, $\{c_{1},...,n_{c}\}$ are the input codes, and, $c_{out}$, the final output $LLC$ code. Sum-pooling is calculated for each dimension of the input codes.

\subsection{Classification\label{sec:svm}}
In this paper, we investigate the performance of state-of-the-art classification algorithms as described in Section \ref{sec:class_perf}. We employed SVM with linear and RBF kernels, Random Forest, Gaussian Process for multi-class classification. We found that SVM with RBF kernel provides the best performance among the surveyed classifiers. In our implementation, LibSVM~\cite{libsvm01} is used and a One-Versus-All strategy is used for multiple-class classification.

\section{Application: Autonomous Clothes Sorting Robot\label{sec:app}}
\begin{figure}[t]
	\centering
	\includegraphics[width= 0.45\textwidth]{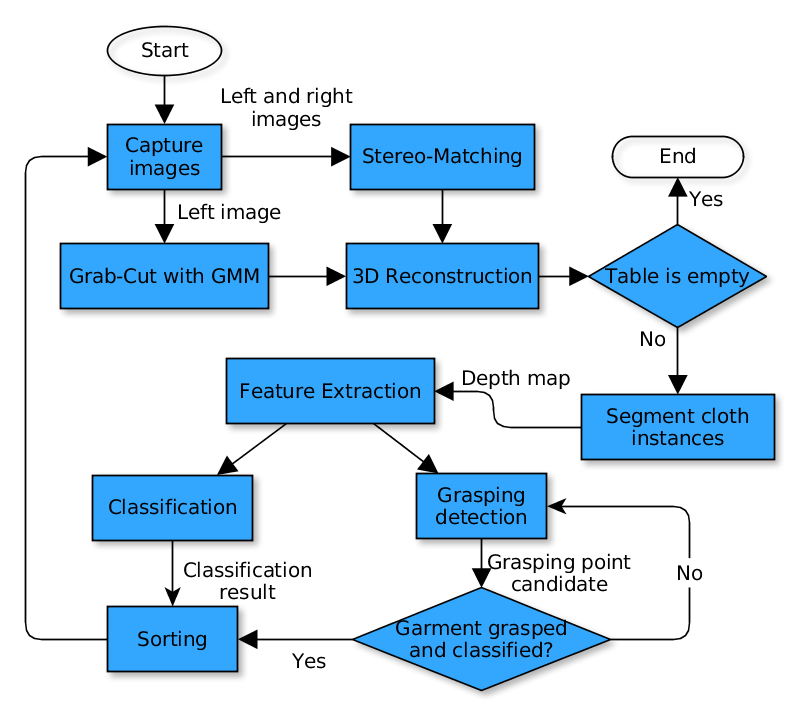}
	\caption{The flow-chart of the autonomous clothes sorting system.}
	\label{fig:flow_chart}
\end{figure}

Clothing category recognition in free-configurations has a range of potential applications, e.g. in a fully-autonomous laundry system. We include our clothes recognition approach into an autonomous robot clothes sorting system (as shown in Fig. \ref{fig:flow_chart}). In this system, a robot categorises each item of clothing in a pile and sorts the clothing into their corresponding box.

In~\cite{lisun2015icra,Aragon-Camarasa2010}, we have described an active binocular robot head system including hand-eye calibration, camera vergence, automatic gaze control, GPU-accelerated stereo matcher and 2.5D reconstruction. As clothes are in free-configuration on a table, segmentation is required to separate clothes from the background. Image segmentation comprise two steps: a supervised grab-cut algorithm~\cite{jan2014garment} is employed to segment the clothing pile from the table. An efficient graph-cut \cite{felzenszwalb2004efficient} is then applied to segment the pile into independent items of clothing. In each sorting iteration, the largest segmented garment in the field of view is selected for recognition and sorting. For each sorting iteration, the status of the table is checked and the largest clothing item is segmented and targeted until the table is empty. 

Following stereo matching, segmentation and 2.5D reconstruction, the depth image is fed into the proposed recognition pipeline to predict the clothing category. Graspable candidates will then be found on the selected item of clothing. A heuristic clothing grasping approach detects and sorts graspable points on the detected wrinkles. For each ridge point, its two contour points are searched for along the two maximum curvature directions on the `ridge' shaped region (similar while detecting wrinkles in Section \ref{sec:outline}). Once a triplet has been matched (a ridge point and its two contour points), it is then treated as a graspable point. After triplet matching, all the graspable positions can be sorted by a flatness ratio obtained from the constructed triplets. During grasping, a success or failure feedback signal is given from the tactile sensor on the tip of the robot gripper \cite{thuy2013development}. In the case of failure, other graspable points are sequentially attempted until the clothing item has been grasped successfully. Once the clothing item has been recognised and grasped successfully, the robot will sort it into the corresponding box. An example of a sorting iteration is shown in Fig.~\ref{fig:sorting}.

\begin{figure*}[t]
	\label{fig:sorting}
	\subfigure[\label{fig:app1}]{\includegraphics[width= 0.16\textwidth]{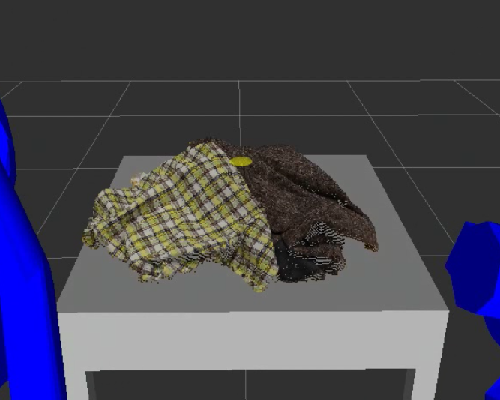}}
	\subfigure[\label{fig:app2}]{\includegraphics[width= 0.16\textwidth]{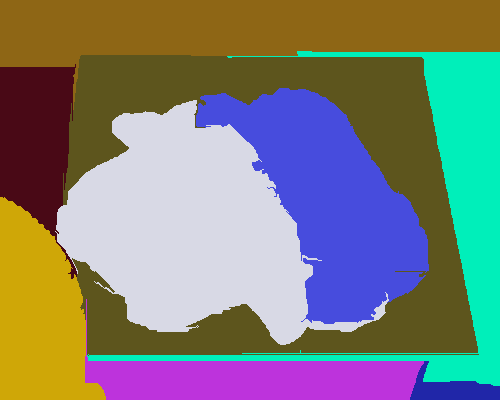}}
	\subfigure[\label{fig:app3}]{\includegraphics[width= 0.16\textwidth]{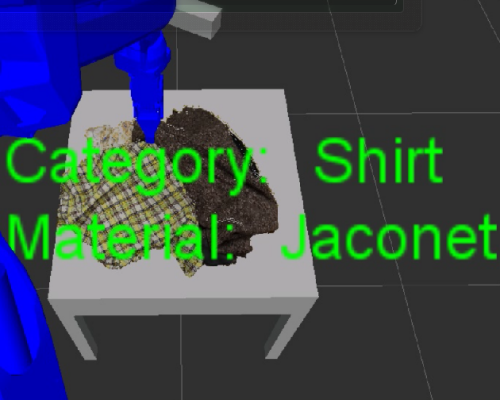}}
	\subfigure[\label{fig:app4}]{\includegraphics[width= 0.16\textwidth]{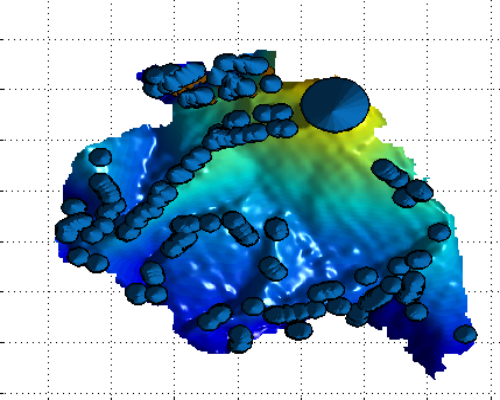}}
	\subfigure[\label{fig:app5}]{\includegraphics[width= 0.16\textwidth]{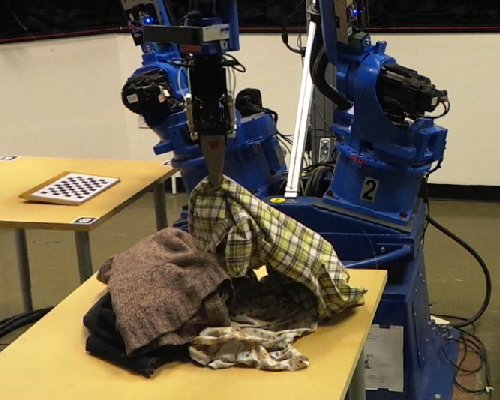}}
	\subfigure[\label{fig:app6}]{\includegraphics[width= 0.16\textwidth]{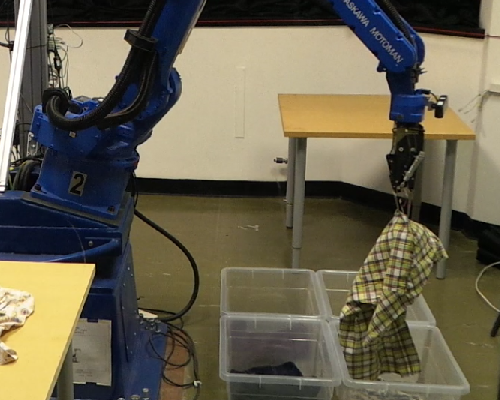}}
	\caption{(a) shows the 2.5D reconstruction as provided by our stereo robot head; (b) is the segmentation result of \cite{jan2014garment}; (c) the robot has classified a garment based on the clothing's material and category; (d) shows estimated grasping points; (d) the garment has been grasped successfully; and, (d) the grasped garment is sorted to its corresponding box.}
\end{figure*}

\section{Experiments\label{sec:experiments}}
We validate our approach in two different scenarios: clothes classification and autonomous robotic sorting experiments. The former measures the performance of recognising categories of clothing items from previously unseen clothing items (Section \ref{sec:clotes_exp}). Whereas, the latter demonstrates our approach in a dual-arm industrial robot test bed (Section \ref{sec:robot_exp}). For the clothes classification experiments, we have captured a high quality RGBD clothing dataset using our stereo head system \cite{lisun2015icra,enlighten72079,aragon2010towards} and the ASUS xtion pro (Section \ref{sec:dataset}). We must note that our approach only employs depth information and RGB data is not included.

The clothes classification validation is further divided into 3 experiments; (a) we fully evaluate the performance of our approach including each local and global features, and the fusion of both feature representations; (b) we investigate the performance of different classification algorithms w.r.t the our visual feature representation; and, (c) we compare the performance between our high-resolution stereo and Asus Xtion sensing devices. As reported in~\cite{willimon2013classification,ramisa2013finddd}, 5-fold cross validation is adopted as the evaluation mechanism of classification performance. We use classification accuracy as our measure of performance. In this paper, we perform the cross validation procedure ten times and average the results.

\subsection{Clothes Dataset\label{sec:dataset}}
Our dataset comprises 50 clothing items of 5 categories: T-shirts, shirts, sweaters, jeans and towels. Material types for each category are: \textit{fine-cotton}, \textit{jaconet}, \textit{wool}, \textit{denim} and \textit{coarse-cotton}, respectively. We have captured 10 clothing items per category of different colours (i.e. from white to black colours). Each item of clothing is captured in 21 different random configurations with our stereo robot head system and the ASUS Xtion.

We have therefore captured 1050 garment samples in random configurations for each sensing device, i.e. 2100 clothing samples. We have provided for each clothing item an RGB image, depth map and segmented mask of 16 MegaPixels (3264$\times$4928) image resolution for our stereo robot head and VGA image resolution for the ASUS Xtion. Our dataset is the first high-resolution free-configuration clothing dataset available. This dataset and our implementation are freely available at: \url{https://sites.google.com/site/clopemaclothesdataset/}. \kevinupdate{In this paper, we resize the images to 4 MegaPixels (1632$\times$2464) as a trade-off between effectiveness and efficiency.}



\subsection{Clothes Classification Experiments\label{sec:clotes_exp}}

\subsubsection{Baseline Performance}
As discussed in Section~\ref{sec:rw}, approaches similar to this paper have been proposed in the literature. Willimon, et.al.'s approach \cite{willimon2013classification} used RGB information of clothes as their visual representation. Hence, their approach is not comparable with our proposed method as our underlying visual representation is based on depth information. Ramisa, et.al.'s approach \cite{ramisa2013finddd}, however, used depth data for feature extraction and classification. We therefore use their implementation as the baseline performance for our approach. In our experiments, we set FINDDD number of bins to $13$, the size of the extraction region without soft voting to $43$ and $85$ for the Asus Xtion and robot head, respectively; and the number of codebooks bases to $512$. These FINDDD parameters provide the best performance in our dataset.

\subsubsection{Feature Representation Performance\label{sec:feat_exp}}

For our local features, $K$-means clustering is employed to create a codebook $B$. For FINDDD and BSP descriptors, we randomly sampled $10^{5}$ descriptors for clustering. For the BSP descriptor (Eq. \ref{eq:4}), the learnt codebook is the reconstructed B-Spline surfaces (Fig. \ref{fig:pipeline}-e). For global features, we use their default parameters (as described in Section \ref{sec:global}) since we did not notice considerable improvement in the accuracy while carrying out our experimental validation (Fig. \ref{fig:pipeline}). As described in Section \ref{sec:class_perf}, SVM with a RBF kernel provides the best classification accuracy, hence, the result herein presented are w.r.t. to this classification algorithm.

For the first group of experiments, the performance of BSP (Section \ref{sec:bsp}), SI (Section \ref{sec:si}), TSD (Section \ref{sec:tsd})and LBP (Section \ref{sec:lbp}) features are evaluated separately. For BSP descriptors, LLC with sum-pooling is employed. The number of codebook bases $K$ is set to 256 \footnote{As shown in Figure \ref{fig:curve}, larger $K$ values correspond to better performance. However, a large $K$ value can potentially over-fit the training data, hence, we employ 256 codebooks as this value depicts a trade-off between over-fitting and generalisation.}. The confusion matrix using BSP+LLC+sum-pooling is shown in Fig. \ref{fig:bsp}. In this figure, diagonal values indicate the accuracy of each category. Rows correspond to the true class, and columns, the predicted class. 
\begin{figure}[t]
\centering
\includegraphics[width= 0.45\textwidth]{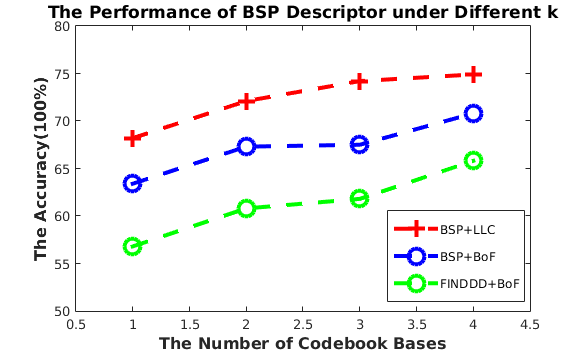}
\caption{The performance of different local descriptors and coding methods. We can observe that BSP with BoF improves the FINDDD descriptor with BoF by approximately 5\%. However, our LLC coding with BSP improves by 10\% w.r.t FINDDD.}
\label{fig:curve}
\end{figure}

\begin{figure*}
	\subfigure[\label{fig:bsp}Local BSP feature $+$ LLC $+$ sum-pooling.]{\includegraphics[width= 0.33\textwidth]{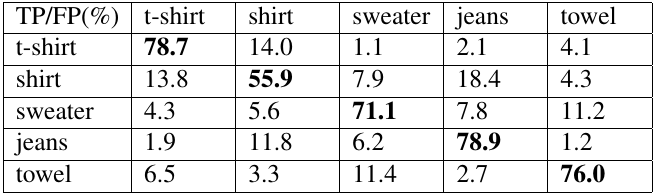}}
	\subfigure[\label{fig:lbp}Global LBP feature.]{\includegraphics[width= 0.33\textwidth]{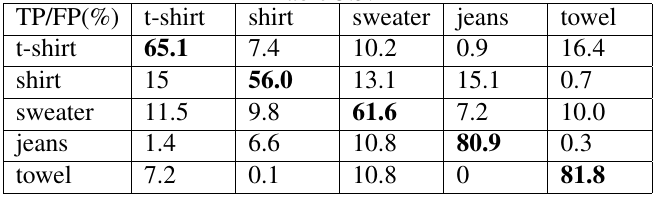}}
	\subfigure[\label{fig:si}Global SI feature.]{\includegraphics[width= 0.33\textwidth]{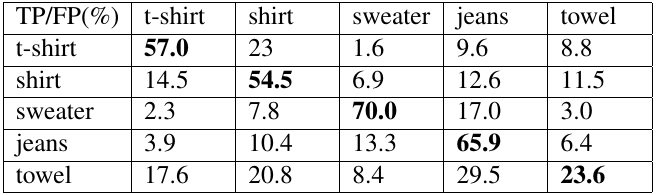}}\\
	\subfigure[\label{fig:tsd}Global TSD feature.]{\includegraphics[width= 0.33\textwidth]{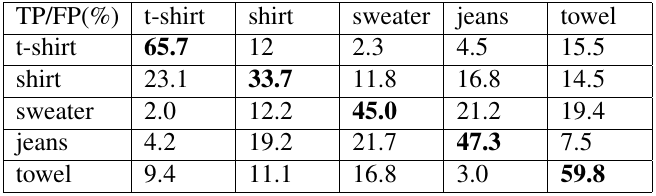}}
	\subfigure[\label{fig:global}Global features (LBP$+$SI$+$TSD).]{\includegraphics[width= 0.33\textwidth]{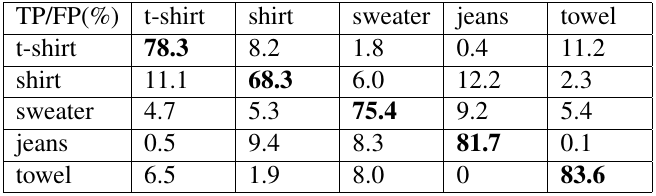}}
	\subfigure[\label{fig:all}Fusion of Local and Global features (L-S-T-B).]{\includegraphics[width= 0.33\textwidth]{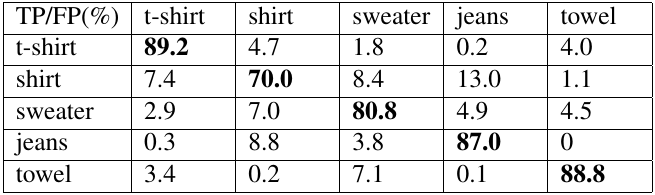}}
	\caption{\label{fig:confmat} Confusion matrices for multi-class  classification. In this figure, the class labels 1-5 correspond to `t-shirts', `shirts', `sweaters', `jeans', `towels'.}
\end{figure*}

As shown in Fig.~\ref{fig:bsp}, \ref{fig:lbp}, \ref{fig:si} and \ref{fig:tsd}, local BSP features achieve a relatively stable performance among all categories. However, the performance of different global features alternate on specific categories. For instance, LBP performs well on jeans and towels, but fails on shirts, while SI works on sweaters but fails on t-shirts and jeans. TSD exhibits the best performance on t-shirts (65.7\% accuracy), but also fails on shirts.

The above suggests that feature fusion will improve the overall performance. To this end, we fuse the three global features (LBP, SI, TSD) with local features (BSP) -- L-S-T-B for short. Experimental results are shown in Fig.{~\ref{fig:global}} and Fig.{~\ref{fig:all}}. We can observe that a fused representation does increase the classification accuracy. Global features achieve 77.5\% accuracy among the $5$ categories, where each category is above 64\%, while L-S-T-B can reach 83.2\% classification accuracy, in which 89.2\% is achieved on t-shirts, 70.0\% on shirts, 80.8\% on sweater, 87.0\% on jeans, and 88.8\% on towels. The classification accuracy of shirts is 16\% lower than the overall average accuracy. \kevinupdate{From our observation, the the configuration of shirts are more susceptible to its various make, thickness, and anti-wrinke property. This likely leads to higher inter-class similarities.}

\subsubsection{Influence on the Classification Algorithm\label{sec:class_perf}}

\begin{table}[t]
\centering
\caption{Table \label{tab:classifier}. Comparison between classification algorithms.}
\begin{tabular}{|p{3.5cm}|p{1.5cm}|}
\hline
Classifiers & Accuracy\%\\
\hline
Random Guess & 20 \\
\hline
FINDDD+BoF+SVM(linear)& 56.2\\
\hline
FINDDD+BoF+SVM(rbf)& \bf{61.8}\\
\hline
L-S-T-B+RF & 72.0  \\
\hline
L-S-T-B+GP(linear) & 79.9 \\
\hline
L-S-T-B+GP(rbf) & 81.0  \\
\hline
L-S-T-B+SVM(linear) & 80.23  \\
\hline
L-S-T-B+SVM(rbf) & \bf{83.2} \\
\hline
\end{tabular}
\end{table}

We investigate the performance of L-S-T-B representation with different classification algorithms. To that end, we evaluate three state-of-the-art classifiers of which Support Vector Machine (SVM), Random Forest (RF) and Gaussian Process (GP) are chosen and implemented for these experiments. For kernel methods (SVM and GP), we analyse the performance of linear and Radial Basis Function (RBF) kernels. For SVM, the cost parameter $C$ is set as $10$, and $\gamma$ of the RBF kernel is set to $10/D$ ($D$ is dimension of L-S-T-D feature description). For RF, we initialise the forest with 2000 trees and $D/5$ dimensions are randomly selected for each tree. We have set these parameters according to our practical experience. GP is couched as a multi-class GP classification problem \cite{gpml2006} where Laplace approximation is used as the inference method. Hyper-parameters of GP kernels are optimised by maximising the log marginal likelihood of the training data.

From Table \ref{tab:classifier}, the performance of RF (72.0\%) is the lowest of all while using our L-S-T-B representation. SVM provides the best classification performance within our pipeline. For both SVM and GP, the performance of RBF kernels are marginally higher than those with linear kernels. Therefore, SVM with RBF kernel is finally selected for our pipeline.

\subsubsection{Depth Sensors Performance}
\begin{table}[t]
\centering
\caption{Table \label{tab:sensor}. Summary of classification accuracy between sensing devices.}
\begin{tabular}{|p{1.7cm}|p{1.6cm}|p{1.6cm}|p{1.6cm}|}
\hline
Accuracy & Asus Xtion & Robot head &  Improvement\\
\hline
Random guess & 20 & 20 & 0\\
\hline
FINDDD+BoF & \bf{47.0\%} & \bf{61.8\%} & \bf{+14.8\%} \\
\hline
BSP+LLC & 52.2\% & 72.1\% & +19.9\% \\
\hline
LBP & 56.4\% & 69.1\% & +12.7\% \\
\hline
SI & 39.1\% & 54.2\% & +15.1\% \\
\hline
TSD & 38.6\% & 50.3\% & +11.7\% \\
\hline
L-S-B & 60.0\% & 77.5\% & +17.5\% \\
\hline
L-S-T-B & \bf{64.2}\% & \bf{83.2}\% & \bf{+19\%}\\
\hline
\end{tabular}
\end{table}

In these experiments, we compare the performance of our proposed clothing category recognition pipeline between sensing devices, i.e. our stereo-head and Asus Xtion. As shown in Table \ref{tab:sensor}, the baseline method (FINDDD+BoF) achieves 47\%. The performance of FINDDD is lower than the reported performance in \cite{ramisa2013finddd}. This is mainly because our experiments are about recognising previously unseen clothes and not about recognising clothing configurations (previously seen clothes) in their paper. In contrast, the accuracy with our binocular robot head is 61.8\% due to the high-quality texture material detail our robot head is able to capture. We can therefore state that our proposed visual representation outperforms previously reported approaches on both Kinect and robot head (64.2\% and 83.2\%, respectively).

\begin{table}[thpb]
\centering
\caption{Table \label{tab:sorting}. Performance of Autonomous Robot Sorting.}
\begin{tabular}{|p{1.3cm}|p{0.8cm}|p{0.7cm}|p{0.7cm}|p{0.7cm}|p{0.7cm}|p{0.7cm}|}
\hline
Categories & T-shirt & Shirt & Sweater & Jeans & Towel & Overall\\
\hline
Success & 7/10 & 4/10 & 7/10 & 7/10 & 8/10 & 33/50 \\
\hline
\end{tabular}
\end{table}

\subsection{Autonomous Robotic Sorting Experiments\label{sec:robot_exp}}

To demonstrate the robustness of our clothing category recognition pipeline in a real-world scenario, we implement our pipeline in an industrial dual-arm robot specifically design to handle and manipulate clothes\footnote{A description of the robotic test bed used in this paper is described in \url{www.clopema.eu/the-robot/}}. We therefore divided 50 clothing items from our dataset into 10 different sorting experiments -- clothing items are only used once for each sorting experiment. Similarly, those selected clothing items for validation are not used for training. This allow us to evaluate the robustness and generalisation of our approach with unseen clothing items. Likewise, sorting performance comprises all modules integrated in our robot: image segmentation \cite{jan2014garment}, stereo-matching and stereo-calibration \cite{lisun2015icra} and robotic manipulation \cite{jan2014garment} as described in Section \ref{sec:app}.

From Table \ref{tab:sorting}, shirts observe low success rate because their intra-class dissimilarity is high. This is consistent to the findings in Section \ref{sec:feat_exp}. Failures of recognition during autonomous sorting are attributed to: (a) the segmentation algorithm fails occasionally when neighbouring clothing items have similar visual appearance; (b) large clothing items are more likely to be affected by occlusions; and, (c) the material properties of soft clothes are reshaped by other rigid clothes while the robot interacts with them. \kevinupdate{Nevertheless, we can therefore argue that our dual-arm robot has the potential to achieve a more efficient autonomous sorting as it requires less perception-manipulation operations (i.e. one-shot clothing item recognition followed by iterative garment grasping) than reported sorting approaches in the literature (e.g. \cite{li2014recognition,willimon2011classification}).} 

\section{Conclusion}
In this paper, we have presented a novel robot vision recognition approach for clothing category in free-configurations based on 2.5D depth information. To the best of our knowledge, our approach is the first to recognise in free-configurations. The classification performance of all proposed features observe a substantial improvement while capturing images with our robot head (Table \ref{tab:sensor}). This is because our robot head can deliver high-quality depth maps which are appropriate for 2.5D surface analysis. Overall, our proposed approach advances the state-of-the-art from 47\% \cite{ramisa2013finddd} to 83.2\% classification accuracy. To demonstrate our approach in a real robotic scenario, we have implemented our approach into an autonomous clothes sorting robot system. In this system, the robot is able to recognise clothing items in a pile, grasp them and sort them into boxes. The whole process is fully autonomous, supported by visual and tactile feedback. Overall, our proposed recognition pipeline therefore advances the state-of-the-art \cite{ramisa2013finddd} in robot clothing recognition.


\addtolength{\textheight}{-5cm}   

\bibliographystyle{IEEEtran}
\bibliography{refs}

\begin{thebibliography}{10}
\providecommand{\url}[1]{#1}
\csname url@rmstyle\endcsname
\providecommand{\newblock}{\relax}
\providecommand{\bibinfo}[2]{#2}
\providecommand\BIBentrySTDinterwordspacing{\spaceskip=0pt\relax}
\providecommand\BIBentryALTinterwordstretchfactor{4}
\providecommand\BIBentryALTinterwordspacing{\spaceskip=\fontdimen2\font plus
\BIBentryALTinterwordstretchfactor\fontdimen3\font minus
  \fontdimen4\font\relax}
\providecommand\BIBforeignlanguage[2]{{%
\expandafter\ifx\csname l@#1\endcsname\relax
\typeout{** WARNING: IEEEtran.bst: No hyphenation pattern has been}%
\typeout{** loaded for the language `#1'. Using the pattern for}%
\typeout{** the default language instead.}%
\else
\language=\csname l@#1\endcsname
\fi
#2}}

\bibitem{marturi2016towards}
N.~Marturi, A.~Rastegarpanah, C.~Takahashi, M.~Adjigble, R.~Stolkin, S.~Zurek,
  M.~Kopicki, M.~Talha, J.~A. Kuo, and Y.~Bekiroglu, ``Towards advanced robotic
  manipulation for nuclear decommissioning: a pilot study on tele-operation and
  autonomy,'' in \emph{Robotics and Automation for Humanitarian Applications
  (RAHA), 2016 International Conference on}.\hskip 1em plus 0.5em minus
  0.4em\relax IEEE, 2016, pp. 1--8.

\bibitem{ramisa2013finddd}
A.~Ramisa, G.~Alenya, F.~Moreno-Noguer, and C.~Torras, ``Finddd: A fast 3d
  descriptor to characterize textiles for robot manipulation,'' in \emph{IROS},
  2013.

\bibitem{willimon2011classification}
B.~Willimon, S.~Birchfleld, and I.~Walker, ``Classification of clothing using
  interactive perception,'' in \emph{ICRA}, 2011.

\bibitem{willimon2013classification}
B.~Willimon, I.~Walker, and S.~Birchfield, ``A new approach to clothing
  classification using mid-level layers,'' in \emph{ICRA}, 2013.

\bibitem{kita2002model}
Y.~Kita and N.~Kita, ``A model-driven method of estimating the state of clothes
  for manipulating it,'' in \emph{IEEE Workshop on Applications of Computer
  Vision (WACV), 2002. Proceedings.}, 2002, pp. 63--69.

\bibitem{kita2009clothes}
Y.~Kita, T.~Ueshiba, E.~S. Neo, and N.~Kita, ``Clothes state recognition using
  3d observed data,'' in \emph{ICRA}, 2009.

\bibitem{kita2009method}
Y.~Kita, T.~Ueshiba, and E.~S. Neo, ``A method for handling a specific part of
  clothing by dual arms,'' in \emph{IROS}, 2009.

\bibitem{li2014recognition}
Y.~Li, C.-F. Chen, and P.~K. Allen, ``Recognition of deformable object category
  and pose,'' in \emph{ICRA, year = {2014}}.

\bibitem{li2014real}
Y.~Li, Y.~Wang, M.~Case, S.-F. Chang, and P.~K. Allen, ``Real-time pose
  estimation of deformable objects using a volumetric approach,'' in
  \emph{IROS}, 2014.

\bibitem{sift2004ijcv}
D.~Lowe, ``Distinctive image features from scale-invariant keypoints,''
  \emph{International Journal of Computer Vision}, vol.~60, no.~2, pp. 91--110,
  2004.

\bibitem{fpfh2009icra}
R.~Rusu, N.~Blodow, and M.~Beetz, ``Fast point feature histograms (fpfh) for 3d
  registration,'' in \emph{ICRA}, 2009.

\bibitem{bof2006}
G.~Csurka, C.~Dance, L.~Fan, J.~Willamowski, and C.~Bray, ``Visual
  categorization with bags of keypoints,'' in \emph{Workshop on statistical
  learning in computer vision, ECCV}, vol.~1, no. 1-22.\hskip 1em plus 0.5em
  minus 0.4em\relax Prague, 2004, pp. 1--2.

\bibitem{lee2006efficient}
H.~Lee, A.~Battle, R.~Raina, and A.~Y. Ng, ``Efficient sparse coding
  algorithms,'' in \emph{Advances in Neural Information Processing Systems},
  2006, pp. 801--808.

\bibitem{wong2014improvements}
T.~Wong, ``Improvements to physically based cloth simulation,'' 2014.

\bibitem{sun2016icra}
L.~Sun, S.~Rogers, G.~Aragon-Camarasa, and J.~P. Siebert, ``Recognising the
  clothing categories from free-configuration using gaussian-process-based
  interactive perception,'' in \emph{ICRA}, 2016.

\bibitem{lbpPAMI2002}
T.~Ojala, M.~Pietikainen, and T.~Maenpaa, ``Multiresolution gray-scale and
  rotation invariant texture classification with local binary patterns,''
  \emph{IEEE Transactions on Pattern Analysis and Machine Intelligence},
  vol.~24, no.~7, pp. 971--987, Jul 2002.

\bibitem{lisun2015icra}
L.~Sun, A.-C. Gerardo, R.~Simon, and J.~P. Siebert, ``Accurate garment surface
  analysis using an active stereo robot head with application to dual-arm
  flattening,'' in \emph{2015 IEEE International Conference on Robotics and
  Automation (ICRA)}, 2015.

\bibitem{lam1992thinning}
L.~Lam, S.-W. Lee, and C.~Y. Suen, ``Thinning methodologies-a comprehensive
  survey,'' \emph{IEEE Transactions on pattern analysis and machine
  intelligence}, vol.~14, no.~9, pp. 869--885, 1992.

\bibitem{koenderink1992surface}
J.~J. Koenderink and A.~J. van Doorn, ``{Surface shape and curvature scales},''
  \emph{Image and vision computing}, vol.~10, no.~8, pp. 557--564, 1992.

\bibitem{vedaldi08vlfeat}
A.~Vedaldi and B.~Fulkerson, ``{VLFeat}: An open and portable library of
  computer vision algorithms,'' \url{http://www.vlfeat.org/}, 2008.

\bibitem{rogers2001introduction}
D.~F. Rogers, \emph{{An introduction to NURBS: with historical
  perspective}}.\hskip 1em plus 0.5em minus 0.4em\relax Morgan Kaufmann, 2001.

\bibitem{wang2010locality}
J.~Wang, J.~Yang, K.~Yu, F.~Lv, T.~Huang, and Y.~Gong, ``Locality-constrained
  linear coding for image classification,'' in \emph{CVPR}, 2010.

\bibitem{libsvm01}
C.-C. Chang and C.-J. Lin, ``{LIBSVM}: A library for support vector machines,''
  \emph{ACM Transactions on Intelligent Systems and Technology}, 2011.

\bibitem{Aragon-Camarasa2010}
G.~Aragon-Camarasa, H.~Fattah, and J.~P. Siebert, ``{Towards a unified visual
  framework in a binocular active robot vision system},'' \emph{Robotics and
  Autonomous Systems}, vol.~58, no.~3, pp. 276--286, Mar. 2010.

\bibitem{jan2014garment}
J.~Stria, D.~Pr\r{u}\v{s}a, V.~Hlav\'{a}\v{c}, L.~Wagner, V.~Petr\'{i}k,
  P.~Krsek, and V.~Smutn\'{y}, ``Garment perception and its folding using a
  dual-arm robot,'' in \emph{IROS}, 2014.

\bibitem{felzenszwalb2004efficient}
P.~F. Felzenszwalb and D.~P. Huttenlocher, ``Efficient graph-based image
  segmentation,'' \emph{International Journal of Computer Vision}, vol.~59,
  no.~2, pp. 167--181, 2004.

\bibitem{thuy2013development}
M.~J. Thuy-Hong-Loan~Le, A.~Landini, M.~Zoppi, D.~Zlatanov, and R.~Molfino,
  ``On the development of a specialized flexible gripper for garment
  handling,'' \emph{Journal of Automation and Control Engineering Vol}, vol.~1,
  no.~3, 2013.

\bibitem{enlighten72079}
W.~Cockshott, S.~Oehler, G.~A. Camarasa, J.~Siebert, and T.~Xu, ``A parallel
  stereo vision algorithm,'' in \emph{Many-Core Applications Research Community
  Symposium 2012}, 2012.

\bibitem{aragon2010towards}
G.~Aragon-Camarasa, H.~Fattah, and J.~Paul~Siebert, ``Towards a unified visual
  framework in a binocular active robot vision system,'' \emph{Robotics and
  Autonomous Systems}, vol.~58, no.~3, pp. 276--286, 2010.

\bibitem{gpml2006}
C.~E. Rasmussen, ``Gaussian processes for machine learning,'' 2006.

\end{thebibliography}

\end{document}